\documentclass{article}

     \usepackage[final]{neurips_2023}

\usepackage{amsmath}
\usepackage{algorithm}
\usepackage{algorithmic}
\usepackage{tikz}
\usepackage{pgfplots}
\usepackage{amsmath}
\usepackage[utf8]{inputenc}
\usepackage{amsmath}
\DeclareUnicodeCharacter{2061}{}
\usepackage[T1]{fontenc}    % use 8-bit T1 fonts
\usepackage{hyperref}       % hyperlinks
\usepackage{url}            % simple URL typesetting
\usepackage{booktabs}   
\usepackage{graphicx}% professional-quality tables
\usepackage{amsfonts}  
\usepackage{amsmath}
\usepackage{nicefrac}       % compact symbols for 1/2, etc.
\usepackage{microtype}      % microtypography
\usepackage{xcolor}         % colors
\usepackage{cite}
\usepackage{natbib}
\usepackage{listings}
\usepackage{xcolor}
\usepackage{soul}
\title{Code Generation and Algorithmic Problem Solving Using Llama 3.1 405B}

\author{%
  Aniket Deroy\\
  Computer Science and Engineering\\
  IIT Kharagpur\\
  \texttt{roydanik18@kgpian.iitkgp.ac.in} \\
   \And
   Subhankar Maity \\
   Department of Artificial Intelligence\\
   IIT Kharagpur \\
   \texttt{subhankar.ai@kgpian.iitkgp.ac.in} \\
}

\begin{document}
\maketitle
\begin{abstract}
Code generation by Llama 3.1 models, such as Meta's Llama 3.1 405B, represents a significant advancement in the field of artificial intelligence, particularly in natural language processing and programming automation. This paper explores the capabilities and applications of Llama-driven code generation, highlighting its ability to translate natural language prompts into executable code across multiple programming languages. Key features include contextual awareness, multi-language support, and enhanced debugging and optimization functionalities. By examining these aspects, we illustrate how Llama can serve as a versatile tool for developers of all skill levels, improving productivity and efficiency in software development. The potential implications for education, industry, and the future of coding practices are also discussed, underscoring the transformative impact of AI in programming. Experimentation shows that while Llama 3.1 405B performs well with simple algorithmic and data structure based problems, it still struggles with problems on Quantum Computing, Bioinformatics, and Artificial Intelligence. 

\end{abstract}

\section{Introduction}

In the evolving landscape of artificial intelligence, one of the most transformative applications is code generation by models like Meta's Llama 3.1 405B \citep{cg12}. Leveraging the power of natural language processing (NLP), these models can bridge the gap between human language and programming languages, offering unprecedented support to developers, educators, and tech enthusiasts \citep{cg9}. Code generation by Llama 3.1 405B allows users to describe their coding needs in plain English and receive functional code snippets or even complete programs in return \citep{cg12, cg11, cg10}. This capability is not limited to a single programming language; it spans a diverse array of languages, including Python, JavaScript, Java, C++, and HTML/CSS. The ability to generate code in multiple languages makes Llama 3.1 405B a versatile tool for a wide range of applications, from web development to data science and beyond. One of the standout features of Llama-driven code generation is its contextual awareness. The model can understand the broader context of a project, ensuring that generated code is consistent and coherent with existing codebases. This contextual understanding extends to debugging and optimization, where Llama 3.1 405B can assist in identifying and rectifying errors or suggesting improvements to enhance performance and efficiency \citep{cg12}. 

In this work, we will explore the key features and benefits of code generation by Llama 3.1 405B, highlighting how this technology can revolutionize the way we write, debug, and optimize code. Whether you are a seasoned developer looking to streamline your workflow or a beginner eager to learn programming, Llama 3.1 405B offers a powerful toolset to enhance your coding experience.
Experimentation with Llama 3.1 405B reveals a strong performance in handling simple algorithmic and data structure-based problems. Tasks such as sorting algorithms, basic search algorithms, and fundamental data structure manipulations are well within its capabilities. The model efficiently processes these problems, providing accurate and timely solutions.

However, when it comes to more specialized and complex fields like Quantum Computing, Bioinformatics, and Artificial Intelligence, Llama 3.1 405B encounters significant challenges. Quantum Computing requires an understanding of quantum mechanics and sophisticated mathematical frameworks, which are beyond the scope of the model's current capabilities. Similarly, Bioinformatics involves the analysis of complex biological data, necessitating knowledge in both biology and data science. The intricacies of genetic sequences, protein structures, and biological pathways present hurdles that the model is not yet equipped to overcome.

In the realm of Artificial Intelligence, the model struggles with advanced concepts such as deep learning architectures, reinforcement learning, and the nuances of neural network training. These areas demand a depth of understanding and specialization that surpasses the model’s current training and architecture. These findings highlight a crucial gap in Llama 3.1 405B's performance, indicating the need for further development and training in these advanced and specialized fields. While the model excels in fundamental computational tasks, its limitations become apparent in the face of cutting-edge scientific and technological challenges. Enhancing its capabilities in these areas is essential for broader applicability and effectiveness in diverse research and application domains.

\section{Related Work}

\noindent \textbf{LLMs for Code Generation and Algorithmic Problem Solving:} Large language models (LLMs) have shown significant promise in the fields of code generation and algorithmic problem solving. Early efforts, such as those by \citep{cg1}, leveraged probabilistic models for generating syntactically and semantically consistent code. The introduction of transformer-based models \citep{cg2} marked a turning point, with OpenAI's GPT-3 \citep{cg3} and Codex \citep{cg4} achieving notable success in generating code snippets and entire programs from natural language prompts. However, challenges in ensuring correctness, efficiency, and handling complex programming constructs persist \citep{cg5, cg11}. AlphaCode \citep{cg6} combined transformers with reinforcement learning to address competitive programming tasks, demonstrating progress but also highlighting the need for deeper reasoning and optimization capabilities in LLMs \citep{cg7}.

\noindent \textbf{LLMs and Transformer based models for Related Tasks:}
With the recent popularity of LLMs, they have been applied for text summarization as well. 
%A work on news document summarization~\cite{zhang2023benchmarking} tried to investigate the performance of multiple Gemini based on the type of prompts and Gemini size. The study found that prompting techniques play a critical role in determining the quality of the summaries. 
%Also, human evaluation with Gemini summaries shows that Gemini summaries have reached the level of human performance for news document. The human-written summaries were high-quality summaries. The human summaries have been written by freelance writers in this study.
A recent work~\citep{10.1145/3551349.3559555} on code summarization using LLMs used zero-shot and few-shot capabilities of LLMs for project-specific code summarization. 
%In the initial stages of a Project there is small amounts of metadata associated with the codes of a project, hence we need to use Gemini like Gemini to find a plausible solution to Code summarization. Gemini show good performance with project-specific Code summarization.
A work on book summarization~\citep{chang2023booookscore} tried to develop novel methods to summarize long books using foundational models like LLMs. They divide the entire book into fragments and then follow a hierarchical strategy of divide and conquer to summarize the entire corpus of a long book.
LLMs have also been used for comment summarization of videos and topic matching for videos~\citep{dimarco2023llm}.
Summarization of long meetings has also been attempted using LLMs like GPT-4 and GPT-3.5 and encoder-decoder models like BART~\citep{schneider2023team}.
There is a work~\citep{maity2024novel} on distractor generation in MCQs where a multi-stage prompting approach delivers better results than a single-stage prompting approach. There are several works on question generation~\citep{maity2023harnessing,maity2024exploring,maity2024effective, aa25} for school-level subjects using GPTs. There are several works on classification~\citep{deroy2021multi,nigam2023nonet}, summarization~\citep{deroy2023prompted,deroy2024ensemble,deroy2023ready,deroy2021analytical,deroy2024applicability,nigam2023fact,deroy2024artificial}, low resource language learning~\citep{maity2024ready}, bias detection in texts~\citep{deroy2023questioning}, temporal reasoning tasks \citep{md1}, mathematical reasoning tasks \citep{qgppf} etc. which use GPTs and Gemini-like LLMs and transformer based models.

In summary, while significant strides have been made in applying LLMs across diverse domains, challenges remain in achieving consistent, reliable performance, especially in complex tasks requiring deep reasoning and optimization. Our investigation into the capabilities of Llama 3.1 405B \citep{cg12} aims to address these gaps, contributing to the broader understanding and enhancement of LLMs in automated programming, algorithmic reasoning, and other critical areas of research.

%The study addresses whether Gemini can be used for human evaluation~\cite{shen2023large} for abstractive summaries. The study shows that Gemini are not very accurate for human evaluation when compared to knowledgable human evaluators. Gemini performing evaluation of abstractive summaries score similar candidate summaries by widely different values thereby questioning the effectiveness of Gemini for human evaluation.
%However, to the best of our knowledge, no work has explored the capabilities of Gemini for mathematical reasoning tasks. In this study, we explore the strengths and limitations of Gemini via mathematical reasoning.

%\section{Methodology}
%We discuss the methodologies behind QG:-
%\subsection{Capabilities in Temporal Reasoning}
%In section 3.1 we will discuss about the capabilities of Gemini models for temporal reasoning.
\section{Dataset}
\label{dataset}
We consulted with experts who are subject teachers in the field of Artificial intelligence (AI), Quantum Computing (QC), Algorithms (AG), Programming and Data Structures (PDS), and Applications on machine learning in Bioinformatics (BioA).
We curated a dataset of 100 programming problems from every subject by consulting with the corresponding subject experts.

\section{Methodology}
We prompted the Llama 3.1 405B model\footnote{\url{https://www.meta.ai/}} to solve problems on every subject mentioned in ~\ref{dataset}. The prompt we provide to the model is: "\textit{Provide a solution to the Problem: <Problem>}".
After obtaining the output from the model we consult the subject expert to find whether the solution provided by the LLM is correct or not correct.

We prompted the GPT-3.5 Turbo model\footnote{\url{https://platform.openai.com/docs/models/gpt-3-5-turbo}} to solve problems on every subject mentioned in ~\ref{dataset}. The prompt we provide to the model is: "\textit{Provide a solution to the Problem: <Problem>}".
After obtaining the output from the model we consult the subject expert to find whether the solution provided by the LLM is correct or not correct.

We prompted the Gemini model\footnote{\url{https://gemini.google.com/app}} to solve problems on every subject mentioned in ~\ref{dataset}. The prompt we provide to the model is: "\textit{Provide a solution to the Problem: <Problem>}".
After obtaining the output from the model we consult the subject expert to find whether the solution provided by the LLM is correct or not correct.

The percentage of correct output is detected by the subject expert after verifying the solution provided by the LLM.

We also performed human evaluation of the generated codes where we asked the subject experts to provide a score to the solution provided by the LLM. We tested two metrics namely - Relevance and Completeness, where we tried to test how relevant and complete is the solution provided by the LLM. We asked three subject experts to provide scores to 50 questions of every category and then the scores were averaged in every category.

\section{Results}

Table ~\ref{program_results1} shows the percentage results of the models on different types of programming problems on different subjects.
The table presents the performance of three different models, Llama 3.1 405B, Gemini, and GPT-3.5 Turbo—across five categories: Algorithms, PDS, AI, BioA, and QC. Each category is scored as a percentage, indicating the performance of each model in that specific area. Llama 3.1 405B shows strong performance in Algorithms (94\%) and PDS (98\%), indicating that it excels in these areas. Its performance in AI is moderate (67\%), while it is relatively lower in BioA (56\%) and QC (54\%). GPT-3.5 Turbo has decent performance in Algorithms (80\%) and PDS (90\%), but its scores are lower than those of Llama-3.1 and Gemini. It shows a significant drop in performance in AI (45\%), BioA (43\%), and QC (40\%), indicating that it is less effective in these areas compared to the other two models. Llama 3.1 405B leads in all categories, especially in Algorithms and PDS. Gemini follows closely behind Llama 3.1 405B, particularly in Algorithms and PDS, but shows a slightly lower performance overall. GPT-3.5 Turbo lags behind the other two models, particularly in AI, BioA, and QC, while maintaining a decent performance in Algorithms and PDS. This comparison highlights the strengths and weaknesses of each model in different categories, helping to identify the most suitable model for specific tasks.

Table ~\ref{program_results} shows the results of the models on different types of programming problems on different subjects for human evaluation. Llama 3.1 405B performs the best in both relevance and completeness, making it the most reliable model among the three for generating highly relevant and thorough responses. Gemini is a strong performer as well, with slightly lower scores than Llama 3.1 405B, but still producing high-quality responses. GPT-3.5 Turbo has the lowest scores in both categories, suggesting that its responses, while still good, may not be as pertinent or complete as those from Llama 3.1 405B or Gemini.

We observe that Llama 3.1 405B performs the best across all LLMs.

\begin{table}
\centering
\caption{Percentage of algorithms written correctly by LLMs as determined by an expert in that subject.}
\small
\begin{tabular}{c|c|c|c|c|c}
\toprule
\textbf{Model} & \textbf{Algorithms} & \textbf{PDS} & \textbf{AI} & \textbf{BioA} & \textbf{QC} \\
\midrule
Llama 3.1 405B & 94\% & 98\% & 67\% & 56\% & 54\% \\
\midrule
Gemini & 86\% & 95\% & 65\% & 52\% & 50\% \\
\midrule
GPT-3.5 Turbo & 80\% & 90\% & 45\% & 43\% & 40\% \\
\bottomrule
%Data 13 & Data 14 & Data 15 & Data 16 & Data 17 & Data 18 \\
%\hline
\end{tabular}
%\caption{Percentage of algorithms written correctly by LLM as determined by expert on that subject}
\label{program_results1}
\end{table}

\begin{table}
\centering
\caption{Human expert evaluation of code generated by LLMs.}
\small
\begin{tabular}{c|c|c}
\toprule
\textbf{Model} & \textbf{Relevance} & \textbf{Completeness}  \\
\midrule
Llama 3.1 405B & 4.84 & 4.43 \\
\midrule
Gemini & 4.63 & 4.25  \\
\midrule
GPT-3.5 Turbo & 4.31 & 4.10 \\
\bottomrule
%Data 13 & Data 14 & Data 15 & Data 16 & Data 17 & Data 18 \\
%\hline
\end{tabular}
%\caption{Percentage of algorithms written correctly by LLM as determined by expert on that subject}
\label{program_results}
\end{table}

\section{Discussion: \textit{Strengths and Flaws in Code Generation and Explanation by Llama-3.1 405B}}

%\subsection{Strengths and Flaws in Code Generation and Explanation by Llama-3.1}

%\begin{enumerate}
    
\subsection{Activity Selection problem}
The activity selection problem is a classic algorithmic problem aimed at selecting the maximum number of non-overlapping activities from a given set. Each activity has a start and end time, and the goal is to choose a subset of activities that don't overlap, maximizing the number of activities chosen. This problem can be efficiently solved using a greedy algorithm, which repeatedly selects the activity that finishes the earliest and then discards all activities that overlap with the selected one. The greedy approach ensures that the optimal solution is found in polynomial time.

\begin{verbatim}
function activitySelection(start, finish):
    n = len(start)
    activities = sorted(zip(start, finish), key=lambda x: x[1])
    
    selected_activities = []
    last_finish_time = 0
    
    for i in range(n):
        if activities[i][0] >= last_finish_time:
            selected_activities.append(activities[i])
            last_finish_time = activities[i][1]
    
    return selected_activities
\end{verbatim}

\noindent \textbf{\textit{\--------------------------- \hl{Subject Expert's Review of Code Generated by Llama 3.1} \------------------------}}

\begin{itemize}
   
\item \textbf{Efficiency:}

\begin{itemize}
   
\item \textit{Time Complexity}: Sorting the activities takes O(nlog⁡n), and iterating through them takes O(n), making the overall time complexity O(nlog⁡n).

\item \textit{Space Complexity}: The space complexity is O(n)*O(n) due to the storage of the activities and the selected activities list.
 
\end{itemize}
\item \textbf{Readability:}

\begin{itemize}
   
\item \textit{Clear Structure}: The code is well-structured, with the sorting step and the selection step clearly separated.

\item \textit{Descriptive Variable Names}: Variables such as start, finish, activities, selected\_activities, and last\_finish\_time are descriptive and self-explanatory.
 
\end{itemize}
\item \textbf{Well-Documented:}

\begin{itemize}
    
\item \textit{Inline Comments}: Although the provided pseudocode does not have inline comments, it would benefit from a few comments explaining key parts of the code, such as the sorting step and the selection criteria.

\item \textit{Function Documentation}: Adding a docstring to the function explaining the parameters and the return value enhances understandability.

\end{itemize}
\item \textbf{Modularization:}

\begin{itemize}
    
\item \textit{Single Responsibility}: The activitySelection function focuses solely on solving the activity selection problem using a greedy approach, adhering to the single responsibility principle.

\item \textit{Reusability}: The function is generic and can be reused with any list of start and finish times, making it versatile.

\end{itemize}
\item \textbf{Correctness:}

\begin{itemize}
   
\item \textit{Accurate Implementation}: The code correctly implements the greedy approach to the activity selection problem by ensuring the proper handling of the activities based on their finish times.

\item \textit{Edge Cases}: The algorithm handles edge cases such as no activities or overlapping activities by correctly implementing the selection criteria.
 
\end{itemize}
 
\end{itemize}

\subsection{KnapSack Problem}
The knapsack problem is a classic optimization problem in computer science and mathematics. Given a set of items, each with a weight and a value, the objective is to determine the number of each item to include in a collection so that the total weight does not exceed a given limit (the capacity of the knapsack) and the total value is as large as possible.
There are different variations of the knapsack problem, including the 0/1 knapsack problem, where each item can either be included or excluded from the knapsack, and the fractional knapsack problem, where fractions of items can be included. The problem is NP-complete, meaning that there is no known efficient solution for solving all instances of the problem in polynomial time.

We asked Llama-3.1 to provide a solution to the KnapSack problem.
%\textbf*{Algorithm}
%\begin{tikzpicture}[
%  item/.style = {rectangle, draw, minimum width=2cm, minimum height=1cm, fill=blue!20},
%  knapsack/.style = {rectangle, draw, minimum width=6cm, minimum height=4cm, fill=gray!20},
%  arrow/.style = {-{Latex[length=3mm, width=2mm]}, thick}
%]

% Items
%\node[item] (item1) {Item 1};
%\node[item, below=of item1] (item2) {Item 2};
%\node[item, below=of item2] (item3) {Item 3};

% Knapsack
%\node[knapsack, right=4cm of item2] (knapsack) {};

% Labels
%\node[above=0.5cm of knapsack] {Knapsack (Capacity: W)};

% Arrows
%\draw[arrow] (item1.east) -- ++(1cm, 0) |- (knapsack.west);
%\draw[arrow] (item2.east) -- (knapsack.west);
%\draw[arrow] (item3.east) -- ++(1cm, 0) |- (knapsack.west);

% Weights and values
%\node[below=0.5cm of item3, align=center] {Each item has:\\ Weight $w_i$ \\ Value $v_i$};

%\end{tikzpicture}

\begin{verbatim}
function knapsack(values, weights, W):
    n = len(values)
    dp = array of size (n + 1) x (W + 1) initialized to 0

    for i from 1 to n:
        for w from 1 to W:
            if weights[i-1] <= w:
                dp[i][w] = max(dp[i-1][w], dp[i-1][w - weights[i-1]] + values[i-1])
            else:
                dp[i][w] = dp[i-1][w]
    
    return dp[n][W]
\end{verbatim}

\noindent \textbf{\textit{\--------------------------- \hl{Subject Expert's Review of Code Generated by Llama 3.1} \------------------------}}

\begin{itemize}
   
\item \textbf{Efficiency:}

\begin{itemize}
    
\item \textit{Time Complexity}: The code operates in O(n*W) time complexity, where n is the number of items and W is the maximum weight capacity. This is optimal for the dynamic programming solution to the knapsack problem.

\item \textit{Space Complexity}: The use of a 2D list dp with dimensions (n+1)*(W+1) results in O(n*W)*O(n*W) space complexity, which is typical for this problem. However, it can be optimized further to O(W)*O(W) by using a 1D list if space is a critical constraint.

\end{itemize}
\item \textbf{Readability:}

\begin{itemize}
    
\item \textit{Clear Structure}: The code is well-organized, with nested loops that clearly indicate the iterative filling of the dp table.

\item \textit{Descriptive Variable Names}: Variables such as values, weights, W, n, and dp are descriptive and self-explanatory, which helps in understanding the code without needing additional comments.

\end{itemize}
\item \textbf{Well-Documented:}
\begin{itemize}
   
\item \textit{Inline Comments}: Though the provided code does not have comments, it would benefit from a few inline comments explaining key parts of the code. This can help other developers or future maintainers understand the logic quickly.

\item \textit{Function Documentation}: Adding a docstring to the function explaining the parameters and the return value enhances understandability.

\end{itemize}
\item \textbf{Modularization:}

\begin{itemize}
    
\item \textit{Single Responsibility}: The knapsack function focuses solely on solving the knapsack problem using dynamic programming, adhering to the single responsibility principle.
\item \textit{Reusability}: The function is generic and can be reused with any list of values and weights, making it versatile.

\end{itemize}

\item \textbf{Correctness:}

\begin{itemize}
   
\item \textit{Accurate Implementation}: The code correctly implements the dynamic programming approach to the knapsack problem by using a recurrence relation and ensuring the proper handling of the weight constraints.
 
\end{itemize}
 
\end{itemize}

\subsection{Quantum Approximate Optimization Algorithm (QAOA)}

The Quantum Approximate Optimization Algorithm (QAOA) is a quantum algorithm designed for solving combinatorial optimization problems. It was introduced by Edward Farhi, Jeffrey Goldstone, and Sam Gutmann. QAOA is particularly notable because it provides a way to leverage quantum computing to find approximate solutions to problems that are classically hard to solve.

To implement the Quantum Approximate Optimization Algorithm (QAOA) in Python, you'll need to use a quantum computing framework such as Qiskit, which is IBM's open-source quantum computing software development kit.

\begin{verbatim}
import numpy as np
import networkx as nx
from qiskit import Aer, execute
from qiskit.circuit import Parameter
from qiskit.circuit.library import RXGate, RZGate, CXGate
from qiskit.optimization.applications.ising import max_cut
from qiskit.optimization.ising.common import sample_most_likely
from qiskit.aqua import QuantumInstance
from qiskit.aqua.algorithms import QAOA
from qiskit.aqua.components.optimizers import COBYLA
from qiskit.visualization import plot_histogram

# Define the problem graph
graph = nx.Graph()
graph.add_edges_from([(0, 1), (0, 2), (1, 2), (1, 3), (2, 3), (3, 4)])

# Get the weight matrix
w = nx.adjacency_matrix(graph).todense()

# Define the qubit operator for the max cut problem
qubit_op, offset = max_cut.get_operator(w)

# Set up QAOA
p = 1  # Number of layers (depth of QAOA)
optimizer = COBYLA(maxiter=100)
qaoa = QAOA(qubit_op, optimizer, p)

# Run the quantum algorithm
backend = Aer.get_backend('qasm_simulator')
quantum_instance = QuantumInstance(backend, shots=1024)
result = qaoa.run(quantum_instance)

# Get the solution
x = sample_most_likely(result.eigenstate)
solution = max_cut.get_graph_solution(x)

# Visualize the results
print('Max-Cut objective value:', result.eigenvalue.real + offset)
print('Solution:', solution)
print('Solution objective:', max_cut.max_cut_value(solution, w))

# Plot histogram of results
plot_histogram(result.eigenstate)

\end{verbatim}

\noindent \textbf{\textit{\--------------------------- \hl{Subject Expert's Review of Code Generated by Llama 3.1} \------------------------}}

\begin{itemize}
    
\item \textit{Optimizer Selection and Parameters}: The optimizer used (COBYLA) might not always converge to the best solution within 100 iterations. It might be beneficial to either increase the number of iterations or consider other optimizers.

\item \textit{Graph Adjacency Matrix Handling}: The adjacency matrix is directly used without converting it to a dense format which is fine for sparse matrices. However, ensure the compatibility with max\_cut.get\_operator.

\item \textit{Objective Value Calculation}: The calculation of the objective value should be carefully handled, especially when interpreting the results from the quantum simulator.

\item \textit{Visualization and Result Interpretation}: The plot\_histogram function is called without explicitly checking if the matplotlib backend is set correctly for plotting in non-interactive environments.\\
 
\end{itemize}

\subsection{Gene Prediction Problem}
Gene prediction is the process of identifying regions of genomic DNA that encode genes. This is a critical step in understanding the functional elements of the genome and in annotating newly sequenced genomes. Gene prediction can be performed using various computational methods and can be broadly categorized into two approaches: ab initio (de novo) prediction and homology-based prediction.

The \texttt{GenePredictor} class encapsulates the functionality for gene prediction, including methods for finding ORFs and translating DNA sequences.

\begin{verbatim}
class GenePredictor:
    def __init__(self, dna_sequence):
        self.dna_sequence = dna_sequence.upper()
        self.start_codon = 'ATG'
        self.stop_codons = ['TAA', 'TAG', 'TGA']
        self.min_gene_length = 100  # Minimum length of a gene

    def find_orfs(self):
        orfs = []
        sequence_length = len(self.dna_sequence)
        
        for frame in range(3):
            orf_start = -1
            for i in range(frame, sequence_length, 3):
                codon = self.dna_sequence[i:i+3]
                if codon == self.start_codon and orf_start == -1:
                    orf_start = i
                elif codon in self.stop_codons and orf_start != -1:
                    if i + 3 - orf_start >= self.min_gene_length:
                        orfs.append((orf_start, i + 3))
                    orf_start = -1

        return orfs

    def translate(self, dna_sequence):
        codon_table = {
            'ATA':'I', 'ATC':'I', 'ATT':'I', 'ATG':'M',
            'ACA':'T', 'ACC':'T', 'ACG':'T', 'ACT':'T',
            'AAC':'N', 'AAT':'N', 'AAA':'K', 'AAG':'K',
            'AGC':'S', 'AGT':'S', 'AGA':'R', 'AGG':'R',
            'CTA':'L', 'CTC':'L', 'CTG':'L', 'CTT':'L',
            'CCA':'P', 'CCC':'P', 'CCG':'P', 'CCT':'P',
            'CAC':'H', 'CAT':'H', 'CAA':'Q', 'CAG':'Q',
            'CGA':'R', 'CGC':'R', 'CGG':'R', 'CGT':'R',
            'GTA':'V', 'GTC':'V', 'GTG':'V', 'GTT':'V',
            'GCA':'A', 'GCC':'A', 'GCG':'A', 'GCT':'A',
            'GAC':'D', 'GAT':'D', 'GAA':'E', 'GAG':'E',
            'GGA':'G', 'GGC':'G', 'GGG':'G', 'GGT':'G',
            'TCA':'S', 'TCC':'S', 'TCG':'S', 'TCT':'S',
            'TTC':'F', 'TTT':'F', 'TTA':'L', 'TTG':'L',
            'TAC':'Y', 'TAT':'Y', 'TAA':'_', 'TAG':'_',
            'TGC':'C', 'TGT':'C', 'TGA':'_', 'TGG':'W',
        }

        protein = ''
        for i in range(0, len(dna_sequence), 3):
            codon = dna_sequence[i:i+3]
            if codon in codon_table:
                protein += codon_table[codon]
            else:
                protein += 'X'  # Unknown amino acid

        return protein

    def predict_genes(self):
        orfs = self.find_orfs()
        genes = []

        for start, end in orfs:
            dna_sequence = self.dna_sequence[start:end]
            protein_sequence = self.translate(dna_sequence)
            genes.append((start, end, dna_sequence, protein_sequence))

        return genes

# Example usage
dna_sequence = "ATGCGTACTGATGCGTACTAGCGTATGCGTAA"
predictor = GenePredictor(dna_sequence)
predicted_genes = predictor.predict_genes()

print("Predicted genes:")
for gene in predicted_genes:
    start, end, dna, protein = gene
    print(f"Start: {start}, End: {end}, DNA: {dna}, Protein: {protein}")
\end{verbatim}

The example (i.e., "Example usage") demonstrates how to use the \texttt{GenePredictor} class to predict genes from a given DNA sequence.
The provided Python code for the gene prediction problem is a simplified and illustrative implementation. 

However, there are several logical and practical flaws that could be addressed to improve the accuracy and robustness of the gene prediction algorithm:

\noindent \textbf{\textit{\--------------------------- \hl{Subject Expert's Review of Code Generated by Llama 3.1} \------------------------}}

\begin{itemize}
    
\item \textit{Start and Stop Codon Identification:}
The code assumes that ORFs only start with "ATG" and end with "TAA", "TAG", or "TGA". However, in reality, there can be multiple start and stop codons, and there might be other regulatory sequences that influence gene boundaries.
It does not account for nested ORFs or overlapping genes.

\item \textit{Handling Frame Shifts:}
The code processes three reading frames separately but does not handle frameshifts within the sequence. Frame shifts can occur due to insertions or deletions in the DNA sequence.

\item \textit{Minimum Gene Length:}
The minimum gene length is hardcoded to 100 nucleotides. This value is arbitrary and may not be appropriate for all datasets or organisms.

\item \textit{Codon Table Coverage:}
The codon table used for translation assumes a simplified set of codons. It does not account for mitochondrial codons or alternative genetic codes that might be present in different organisms.

\item \textit{Error Handling:}
The translate method appends 'X' for unknown amino acids, but this approach can lead to misinterpretation of the protein sequence. It would be better to handle or log such cases explicitly.

\item \textit{Data Preprocessing:}
The code does not include any data preprocessing steps, such as handling lower-case letters, ambiguous nucleotides (e.g., 'N'), or sequencing errors.

\item \textit{Efficiency:}
The code processes each frame and codon sequentially, which can be inefficient for long sequences. Optimizations, such as using numpy arrays or parallel processing, could improve performance.

\end{itemize}

\subsection{A* algorithm}
The A* (A-star) algorithm is a popular and efficient pathfinding and graph traversal algorithm. It is used to find the shortest path between two nodes in a weighted graph, such as in navigation systems, game development, and artificial intelligence applications.

We discuss the A* algorithm generated by Llama-3.1
\begin{algorithm}[H]
%\scriptsize
\caption{A* Algorithm}
\begin{algorithmic}[1]
\STATE \textbf{Input:} Graph $G = (V, E)$, start node $start$, goal node $goal$, heuristic function $h$
\STATE \textbf{Output:} Path from $start$ to $goal$

\STATE $openSet \gets \{start\}$
\STATE $cameFrom \gets \text{empty map}$
\STATE $gScore \gets \text{map with default value } \infty$
\STATE $gScore[start] \gets 0$
\STATE $fScore \gets \text{map with default value } \infty$
\STATE $fScore[start] \gets h(start)$

\WHILE{$openSet \neq \emptyset$}
    \STATE $current \gets \text{node in } openSet \text{ with lowest } fScore$
    \IF{$current = goal$}
        \STATE \textbf{return} \text{ReconstructPath}$(cameFrom, current)$
    \ENDIF
    \STATE $openSet \gets openSet \setminus \{current\}$
    \FOR{each neighbor $neighbor$ of $current$}
        \STATE $tentative\_gScore \gets gScore[current] + \text{dist}(current, neighbor)$
        \IF{$tentative\_gScore < gScore[neighbor]$}
            \STATE $cameFrom[neighbor] \gets current$
            \STATE $gScore[neighbor] \gets tentative\_gScore$
            \STATE $fScore[neighbor] \gets gScore[neighbor] + h(neighbor)$
            \IF{$neighbor \notin openSet$}
                \STATE $openSet \gets openSet \cup \{neighbor\}$
            \ENDIF
        \ENDIF
    \ENDFOR
\ENDWHILE
\STATE \textbf{return} \text{failure}
\end{algorithmic}
\end{algorithm}

\textbf{Heuristic Function}
The heuristic function $h(n)$ estimates the cost to reach the goal from node $n$. It should be admissible, meaning it never overestimates the actual cost.

\textbf{Example}
Consider a simple grid with the start at $(0, 0)$ and the goal at $(2, 2)$. The heuristic function $h$ can be the Manhattan distance:

\[
h(x, y) = |x_{goal} - x| + |y_{goal} - y|
\]

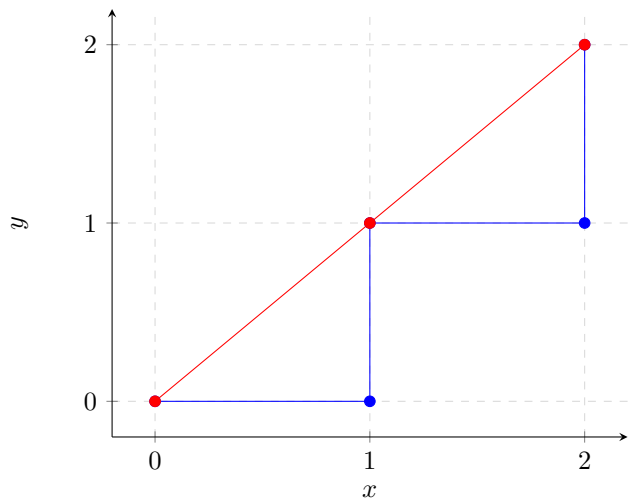
\begin{tikzpicture}
    \begin{axis}[
        axis lines = left,
        xlabel = $x$,
        ylabel = $y$,
        xtick={0, 1, 2},
        ytick={0, 1, 2},
        enlargelimits=true,
        grid = major,
        grid style={dashed, gray!30},
    ]
    \addplot[mark=*, color=blue] coordinates {(0,0) (1,0) (1,1) (2,1) (2,2)};
    \addplot[mark=*, color=red] coordinates {(0,0) (1,1) (2,2)};
    \end{axis}
\end{tikzpicture}

The blue path represents the path found by the A* algorithm.

\noindent \textbf{\textit{\--------------------------- \hl{Subject Expert's Review of Code Generated by Llama 3.1 }\------------------------}}

\begin{itemize}
   
\item \textit{Selecting current node}: The line no 10 in the algorithm implies iterating through openSet to find the node with the lowest fScore. This can be inefficient because it requires O(n)*O(n) time to find the minimum. Using a priority queue (min-heap) would make this operation more efficient, reducing it to O(log⁡n)*O(logn).

%openSet Handling:

%    The removal of current from openSet using openSet \gets openSet \setminus \{current\} is correct but can be inefficient if openSet is implemented as a basic set or list. Using a priority queue also addresses this issue.

\item \textit{Edge Case Handling}: The algorithm does not explicitly handle the case where the start node is the goal node. While it would eventually return the correct result, an immediate check could make it more efficient.

\item \textit{Consistency Check}: While the algorithm correctly updates cameFrom, gScore, and fScore, it doesn't explicitly check for consistency of the heuristic function. Although not a flaw in the implementation itself, a note or comment about ensuring heuristic consistency would be useful.

\item \textit{Heuristic Validity}: The algorithm assumes the heuristic function h is admissible and consistent but doesn't enforce or check this. Adding comments or documentation about this requirement would be beneficial.
 
\end{itemize}
\section{Conclusion and Future Directions}

LLMs generate code based on patterns in the data they were trained on, but they lack true understanding of the context. This can lead to code that superficially looks correct but fails to meet the specific requirements or constraints of the problem. Generated code can contain syntax errors or semantic mistakes that prevent it from running correctly. While LLMs are good at producing syntactically correct code, they can still make errors that a human programmer would avoid. While Large Language Models (LLMs) like GPT-3.5 and GPT-4 can generate code that functions correctly, the code is often not optimized for performance, efficiency, or readability. This lack of optimization arises from several inherent limitations of LLMs

The advent of GPT-driven code generation marks a pivotal moment in the intersection of artificial intelligence and software development. By leveraging the capabilities of models like GPT-4, developers can translate natural language descriptions into functional code, significantly streamlining the coding process. This technology enhances productivity, supports multi-language programming, and offers sophisticated debugging and optimization tools. The result is a powerful, user-friendly interface that democratizes coding, making it more accessible to both novice programmers and seasoned developers. GPT's ability to understand and generate contextually relevant code snippets can reduce the time spent on repetitive coding tasks and minimize human errors. As a result, developers can focus more on creative and complex aspects of software development. Furthermore, the integration of AI in coding workflows has the potential to foster innovation and accelerate the development of new applications, ultimately driving progress in various technological domains.
Experimentation shows that while Llama 3.1 405B performs well with simple algorithmic and data structure-based problems, it still struggles with more complex and specialized areas such as Quantum Computing, Bioinformatics, and Artificial Intelligence. Despite its robust performance in foundational topics, it faces challenges when addressing the intricacies and advanced concepts inherent in these specialized fields. This suggests that while the model is adept at handling basic computational problems, further enhancements are needed for it to effectively tackle the complexities of cutting-edge scientific and technological disciplines.
%\subsubsection{}
\bibliography{references}{}
\bibliographystyle{plainnat}

\end{document}